\documentclass[conference]{IEEEtran}
\IEEEoverridecommandlockouts
\usepackage{cite}
\usepackage{amsmath,amssymb,amsfonts}
\usepackage{algorithmic}
\usepackage{graphicx}
\usepackage{textcomp}
\usepackage{xcolor}
\def\BibTeX{{\rm B\kern-.05em{\sc i\kern-.025em b}\kern-.08em
    T\kern-.1667em\lower.7ex\hbox{E}\kern-.125emX}}
\begin{document}

\title{Towards Trustworthy Keylogger detection: A Comprehensive Analysis of Ensemble Techniques and Feature Selections through Explainable AI\\

\thanks{Identify applicable funding agency here. If none, delete this.}
}

\author{\IEEEauthorblockN{Monirul Islam Mahmud}
\IEEEauthorblockA{\textit{Dept. of Computer \& Information Science} \\
\textit{Fordham University}\\
New York, United States\\
mim9@fordham.edu \\
ORCID: 0009-0005-2812-8553}

}

\maketitle

\begin{abstract}
Keylogger detection involves monitoring for unusual system behaviors such as delays between typing and character display, analyzing network traffic patterns for data exfiltration. In this study, we provide a comprehensive analysis for keylogger detection with traditional machine learning models (SVC, Random Forest, Decision Tree, XGBoost, AdaBoost, Logistic Regression and Naive Bayes) and advanced ensemble methods including Stacking, Blending and Voting. Moreover, feature selection approaches such as Information gain, Lasso L1 and Fisher Score are thoroughly assessed to improve predictive performance and lower computational complexity. The Keylogger Detection dataset [1] from publicly available Kaggle website is used in this project. In addition to accuracy-based classification, this study implements the approach for model interpretation using Explainable AI (XAI) techniques — namely SHAP (Global) and LIME (Local) — to deliver finer explanations for how much each feature contributes in assisting or hindering the detection process. To evaluate the models result, we have used AUC score, sensitivity, Specificity, Accuracy and F1 score. The best performance was achieved by AdaBoost with 99.76\% accuracy, F1 score of 0.99, 100\% precision, 98.6\% recall, 1.0 specificity and 0.99 of AUC that is near-perfect classification with Fisher Score. Compared to model decision transparency provided by other explainable AI methods, results were further validated to establish predictive robustness using ROC curves and confusion matrices, which form an ideal combination of a predictive and defensive tool against keylogger threats.
\end{abstract}


\section{Introduction}

One type of malware termed as keylogger is used to record user keystroke attempts. Cybercriminals typically use keyloggers to watch the target's keystrokes and steal data from their computer or other computing devices. In general, there are two types of keyloggers. The first kind of keyloggers is inserted into the user device to steal the information through the keystrokes. In that instance, the primary focus of hackers was on acquiring the login credentials, such as user IDs and passwords, that users enter to access any website or application. The keyloggers capture each keystroke and transmit it to the fraudsters. Cybercriminals utilize the keylogger in this procedure to obtain private data. Malware often operates in secret, and the user won't realize they have an infection until the computer has been harmed [2]. Cybercriminals can use ransomware to encrypt a computer and demand money to unlock it. If the user disagrees with their specifications, they have the power to delete the data or harm the hardware [3]. Programs known as keyloggers are placed on computers and record each keystroke. As a result, the numerous websites a user visits provide credentials, including usernames and passwords, to hackers. It is advised to avoid using public devices for sensitive transactions or to enter usernames and passwords because there is a genuine chance that keyloggers could be installed on them [4]. 

Machine learning (ML) has a great potential in malware and keylogger detection, since it can learn complicated patterns from data, perform detection automatically. Yet, ML-based cybersecurity solutions are still challenged by the following aspects: High dimensionality where raw system or behavioral datasets typically contain hundreds of features, many of which are irrelevant or redundant. This can lead to poor model performance and a more expensive need for computation. Many real-world settings feature a larger number of benign than malicious instances requiring models to be careful that they are not biasing to predicting the majority class. Especially the ensemble and deep learning based high-performing models, are black-boxes. Due to this, they are unable to be trusted in essential ecosystems like that of finance, healthcare and government systems. To fill the research gaps,  our proposed novel approaches are given below:

\begin{itemize}
\item In this article, we outline an ML pipeline for detecting malware and keyloggers that places equal focus on high explanation and high accuracy to solve these problems. We used  machine learning models and advanced ensemble models including stacking, blending and voting in this research work.

\item Feature selection approaches such as information gain, Information gain, Lasso L1 and Fisher score are assessed to improve predictive performance and lower computational complexity.

\item  We also integrated SHapley Additive exPlanations (SHAP) and Local Interpretable Model-agnostic Explanations (LIME) to increase confidence in the predictions made by the model. They provide a way to understand how the individual features contribute to the predictions and how a classifying a program as a malicious program or a benign program can be explained to the stakeholders.

\item Lastly, balanced detection performance with computational efficiency and interpretability, which offers a deployable solution for real-world systems.

\end{itemize}

\section{Related Works}

Keyloggers are incredibly harmful tools that monitor all of our computer activity.  Every key the user pushes can be recorded by the keylogger, which can then store the data in a log file and email the file to the designated IP address. The financial system, which is used for everyday business operations, is seriously at risk from it. The types, functions, and characteristics of several keyloggers are described in detail [5].  Pillai showed how to detect keyloggers installed on a computer using a modified SVM-based architecture. Eight open source keyloggers were installed on their system [6]. Brown introduced well-known techniques for Android keylogger classification, including XOR, GEFeS, and SDM [16]. Wen L. et al. introduced the unsupervised dimensionality reduction approach and the supervised learning classifier SVM and PCA-RELIEF [7]. A novel data-collecting method based on a unified activity list was used to obtain the novel dataset shown in [8], which was collected in a realistic environment.  The overall average accuracy was 79\% for the binary class, which has two target variables, and 77\% for the multi-class, which has more than two target variables. The breadth of the monitoring included all dataset aspects, including keyloggers, OS activity, microphone access, phone calls, and social media access.  According to the findings of [9], random forest performs better in malware detection than deep neural network models. Random Forest attained the maximum accuracy at 99.78\%. Using four classifiers—ID3, K-Nearest Neighbors, Decision Tree, C4.5 Decision Tree, and linear SVM, the authors of [10] demonstrated a supervised classification method for identifying Android malware (SVM).  

Alghamdi et al. A model based on SVM, Random Forest (RF) and Decision Trees to detect keyloggers and spyware using machine learning is proposed in [11]. The RF classifier obtained 99.6\% accuracy which shows the effectiveness of ML techniques in identifying malicious behavior based on keystrokes and patterns [14. An AI-based method that integrates different artificial intelligence techniques to monitor the system behavior to characterize implicit keys and user input patterns by analyzing their keylogger activity, was proposed by Levshun [12]. Their technique focuses more on detecting behavior rather than relying on old signature-based detection, enabling the identification of new, unidentified keyloggers. Venkatesh et al. They proposed a multi-layered approach using a Random Forest Classifier to identify potentially abnormal keylogging behaviour [13]. Furthermore, Fu et al. For instance, [14] used digital immune system inspired algorithm called Dendritic Cell Algorithm (DCA) in the detection of software keyloggers. The DCA utilizes signals and contextual information to discriminate between expected and malicious behaviors, providing a fresh viewpoint on anomaly detection in the field of cybersecurity.

\section{Methodology}

In this study, we propose a comprehensive methodology for detecting keyloggers using machine learning techniques. These include data acquisition, data preprocessing, feature selection, Machine learning models and Exaplianble AI analysis. For this purpose, we used a dataset called the “Keylogger Detection” [1] from Kaggle, consisting of system process data characterized as benign or malicious. 

\begin{figure}[!htpb]
    \centering
    \includegraphics[width=\linewidth]{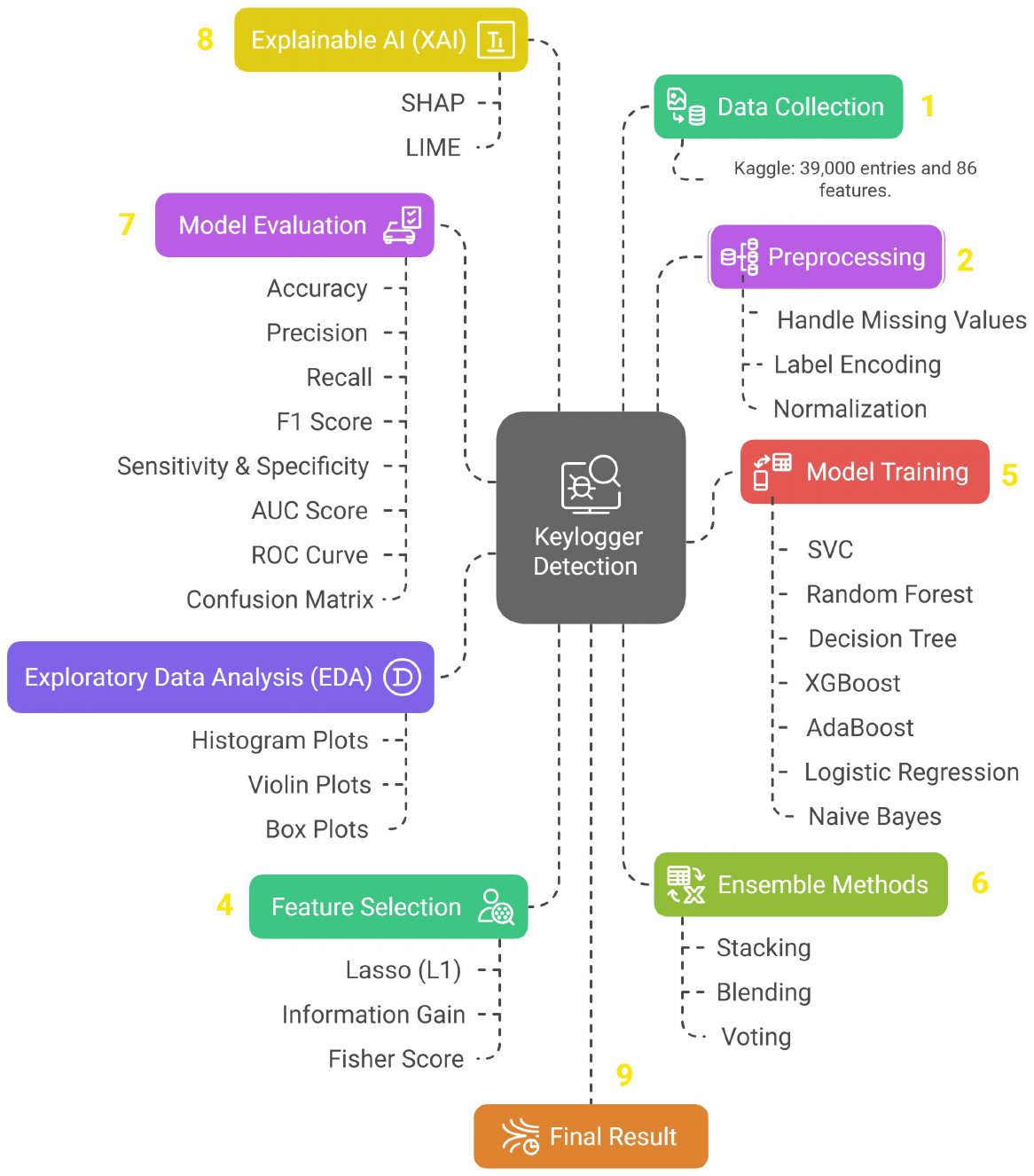}
    \caption{Overall methodology.}
    \label{fig:tcanther}
\end{figure}

\subsection{Data Collection and Preprocessing}

This dataset is collected from Kaggle and it consists of 39,000 network traffic instances and 86 features. We removed columns in the dataset like "Unnamed: 0", "Flow ID", "Timestamp", "Source IP", "Destination IP", since these attributes are not important to keylogger behavior. It has no null values so None need to be imputed. For exploratory data analysis, we plotted correlation matrices to check whether there was multicollinearity between features, histograms for feature distributions, and box and violin plots for outlier detection.
\begin{figure}[!htpb]
    \centering
    \includegraphics[width=\linewidth]{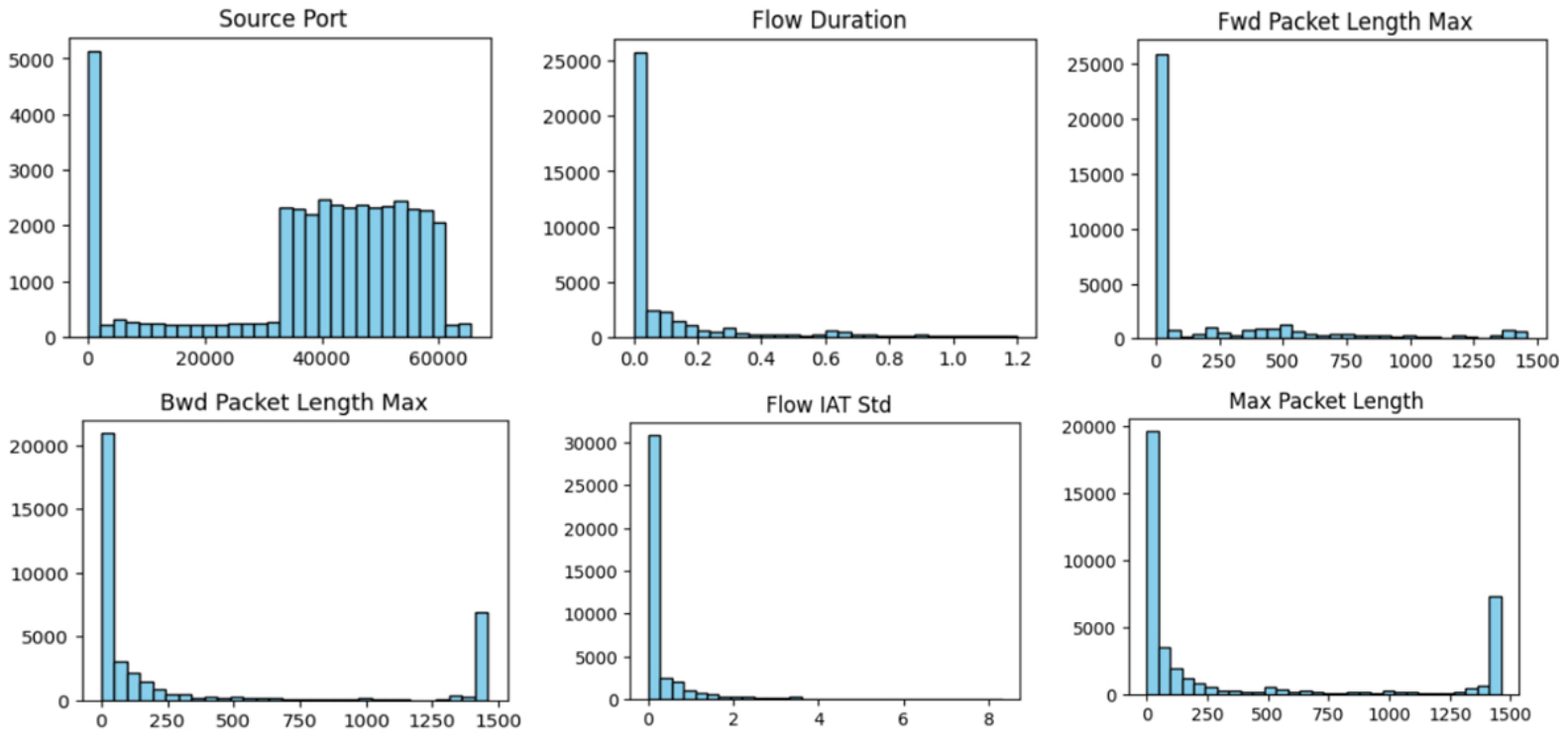}
    \caption{Barplot Distribution.}
    \label{fig:tcanther}
\end{figure}
\begin{figure}[!htpb]
    \centering
    \includegraphics[width=\linewidth]{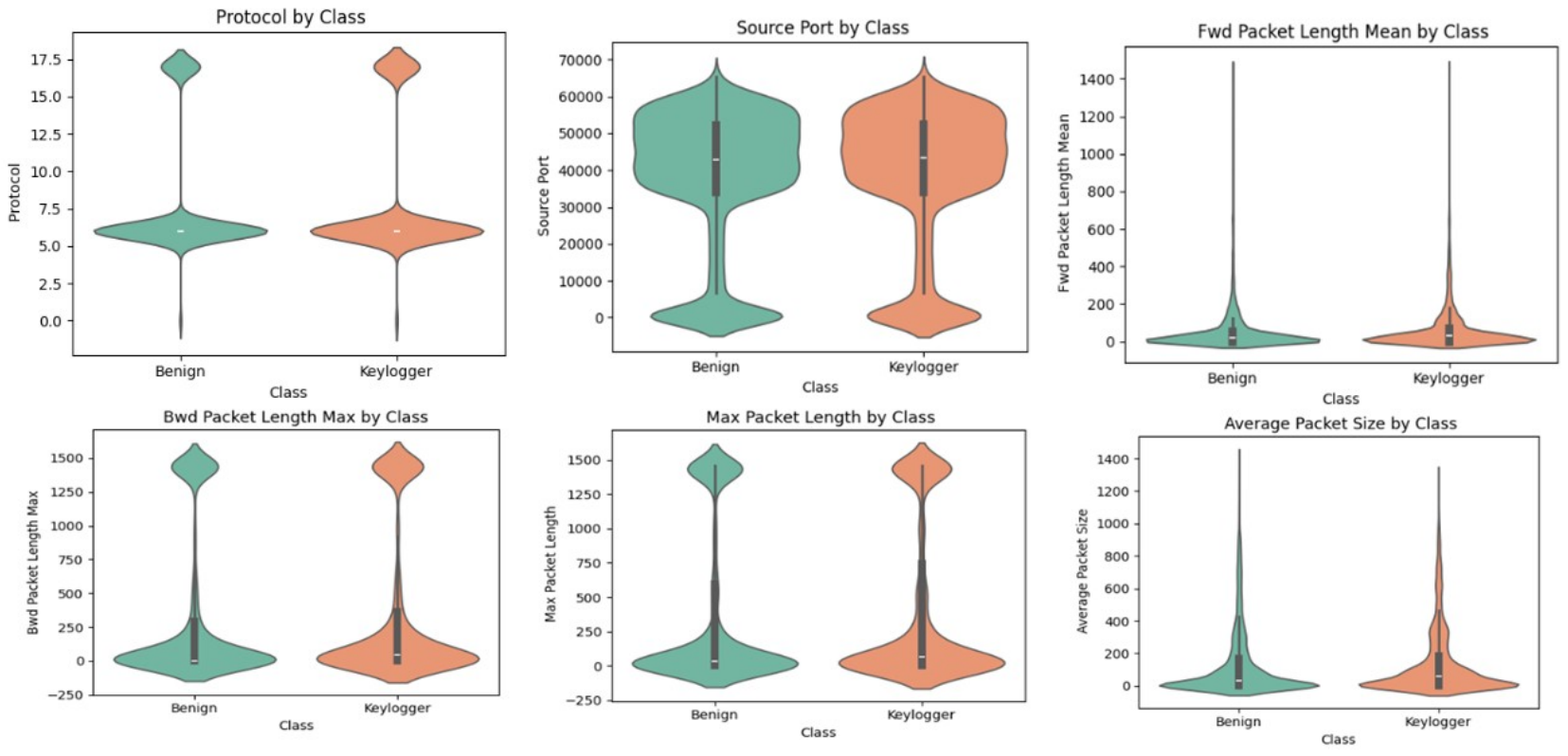}
    \caption{Violin Plot Distribution.}
    \label{fig:tcanther}
\end{figure}
\begin{figure}[!htpb]
    \centering
    \includegraphics[width=\linewidth]{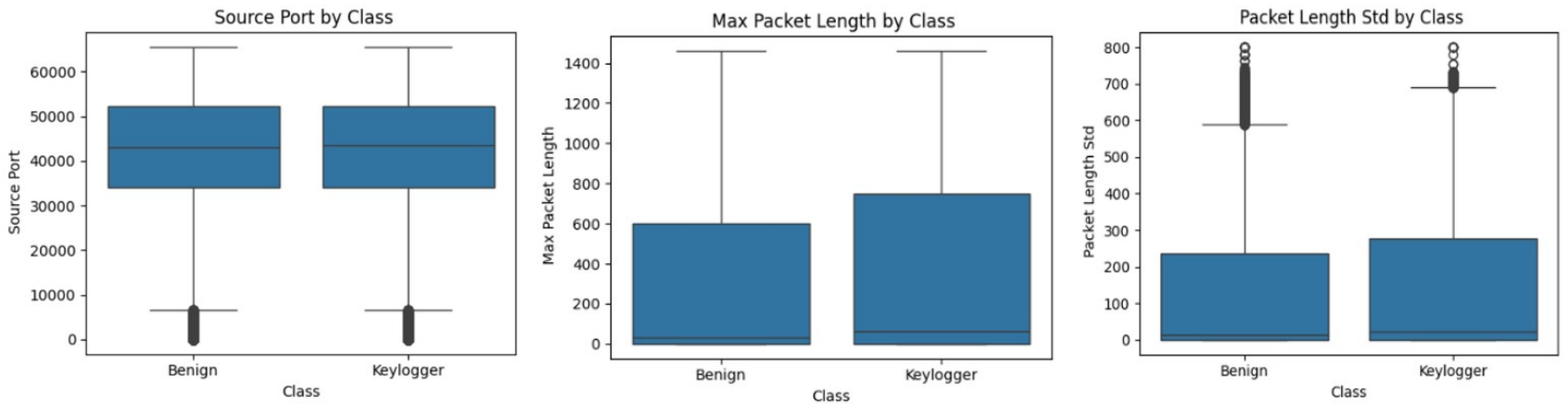}
    \caption{Boxplot Distribution.}
    \label{fig:tcanther}
\end{figure}
The class imbalance was handled using the Synthetic Minority Oversampling Technique (SMOTE), such that both keylogger and benign samples are well represented. Label encoding was performed to convert categorical features to numerical class column labels, And MinMax Scaling was applied to normalize all numerical features to the range [0, 1]. To maintain the integrity of evaluation, the dataset was divided into training and testing sets with 80\% and 20\% respectively.
\begin{figure}[!htpb]
    \centering
    \includegraphics[width=\linewidth]{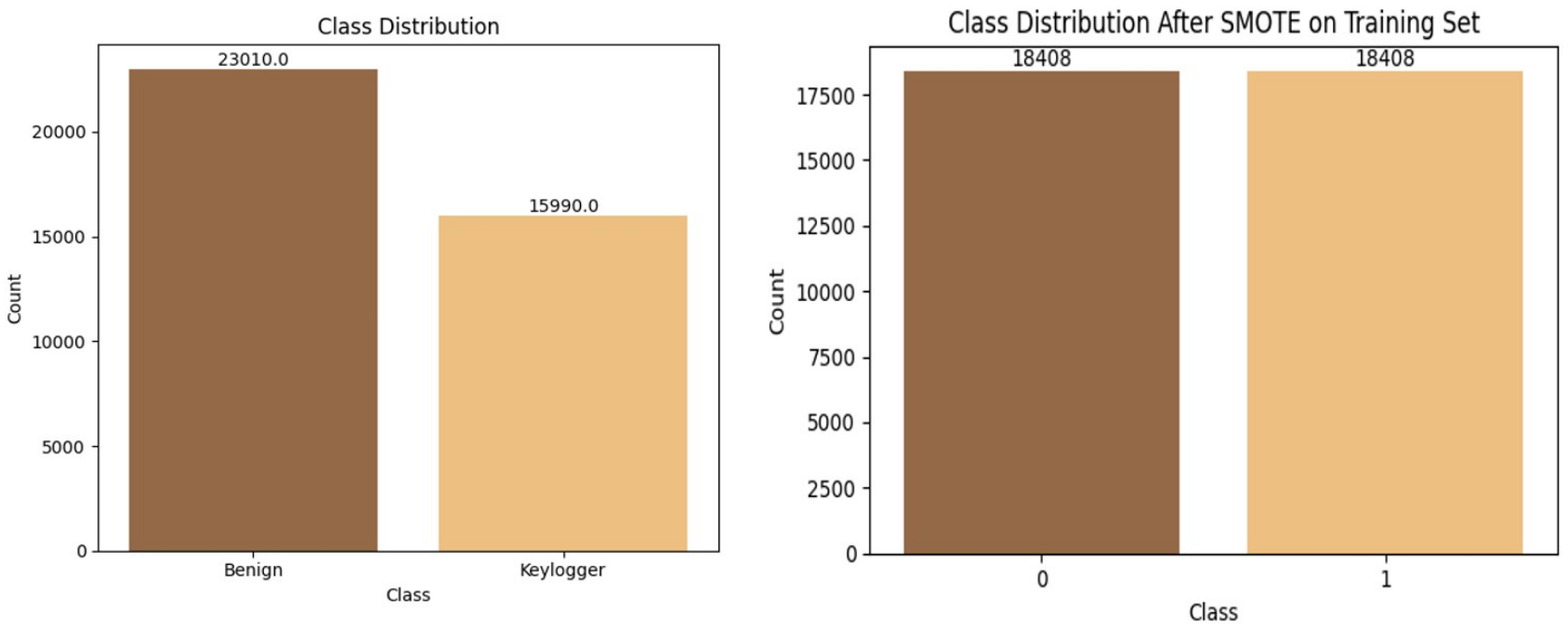}
    \caption{Target class distribution before and after SMOTE.}
    \label{fig:tcanther}
\end{figure}

\subsection{Feature Selection}

Information Gain (IG)–filter-based feature selection method IG [66] quantifies the dependence between each feature and the target variable. It quantifies the change in entropy (i.e. uncertainty) about the target class given that we observe a feature. Higher IG value indicates more information of the feature in distinguishing keylogger traffic from the required benign network activity. To keep the most informative features, threshold (IG higher than 0.1) was applied. This attenuation left 46 features (i.e., Dst\_Port, Pkt\_Len\_Std) that were trained with the model while 40 low-scoring features were eliminated.
\begin{figure}[!htpb]
    \centering
    \includegraphics[width=\linewidth]{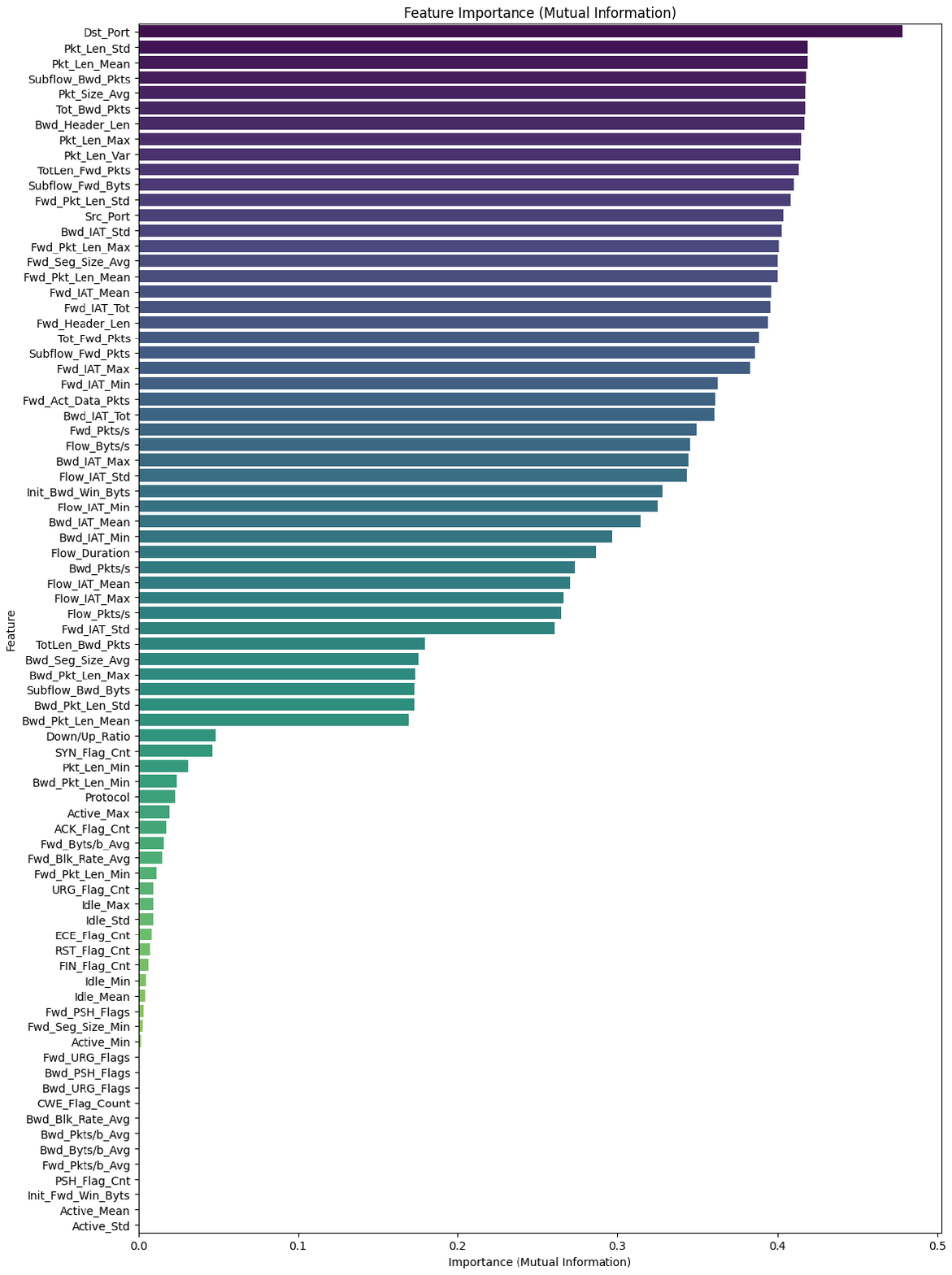}
    \caption{Information Gain distribution.}
    \label{fig:tcanther}
\end{figure}
Lasso (Least Absolute Shrinkage and Selection Operator) is an embedded feature selection method that applies L1 regularization. It punishes the absolute size of coefficients and pushes less important features to zero. Of the 86 candidate features, the L1 penalty selected 56 with nonzero coefficients, discarding 30 redundant or noisy features.
\begin{figure}[!htpb]
    \centering
    \includegraphics[width=\linewidth]{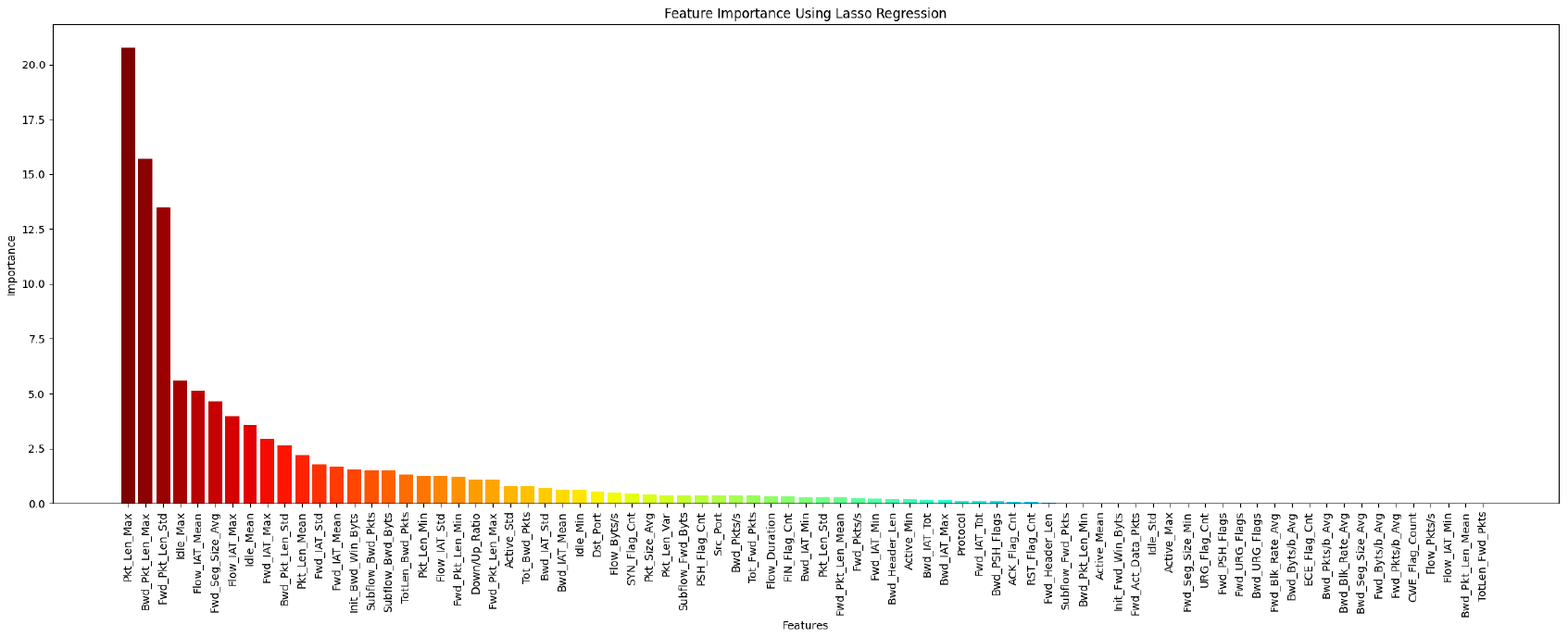}
    \caption{Lasso L1 distribution.}
    \label{fig:tcanther}
\end{figure}
Fisher Score  is one of the computed methods that find features that maximize inter-class (keylogger vs. benign) separation whilst ensuring that intra-class variance is minimized. Discriminating features have a higher Fisher Score. Finally, we selected the top 47 features with the greatest scores. For example, Fisher Score tends to work outperform when classes are well separated as in keylogger detection. As shown, AdaBoost achieved accuracy better than IG and Lasso (99.75\% with Fisher score vs. 99.51\% with IG and 99.02\% with Lasso).
\begin{figure}[!htpb]
    \centering
    \includegraphics[width=\linewidth]{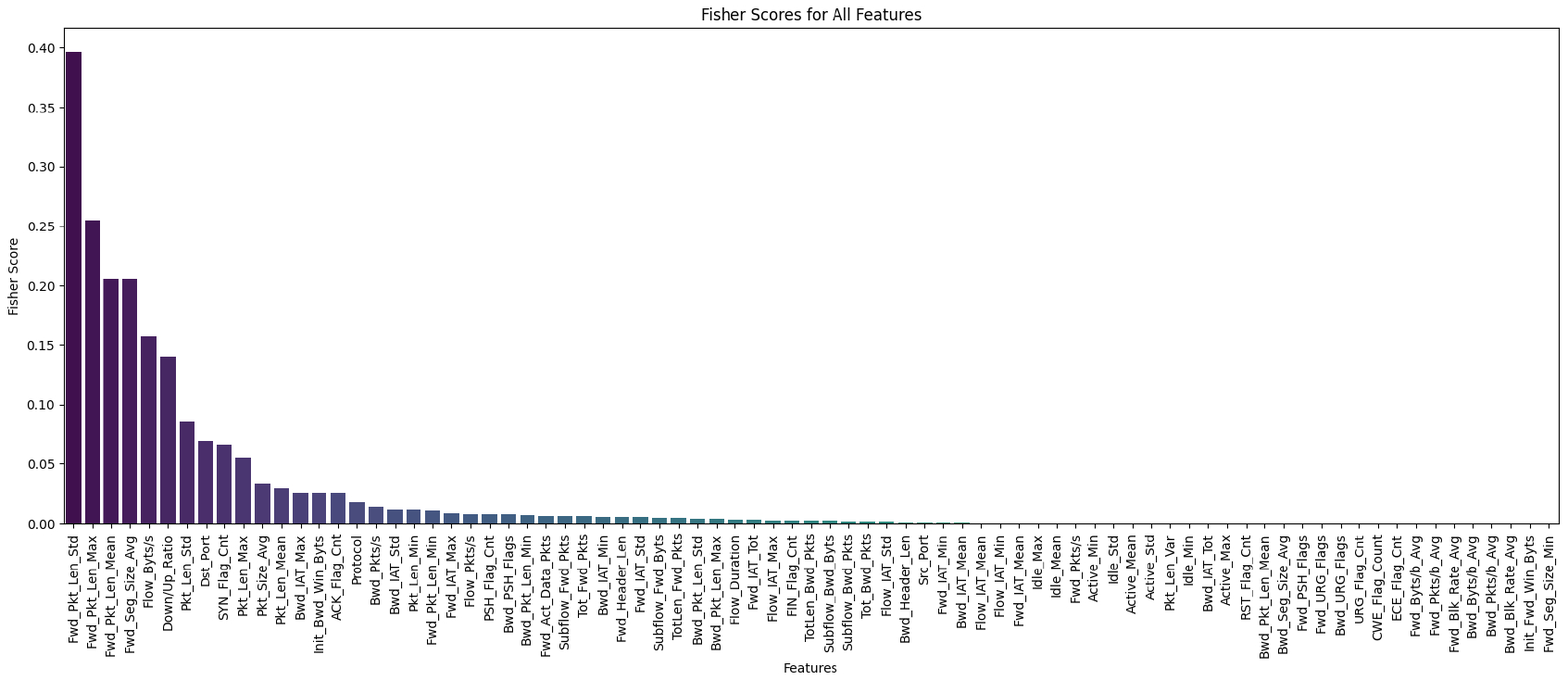}
    \caption{Fisher Score distribution.}
    \label{fig:tcanther}
\end{figure}

\subsection{Model Development}

Nonetheless, we use a comprehensive set of 10 machine learning algorithms to evaluate our models, allowing us to explore different ways to formulate the keylogger detection problem. We used the RBF kernel version of SVC, which is very effective with non-linear decision boundaries and high-dimensional feature space. This makes random forst extremely useful when the interpreting feature importances are important and avoids the problem of overfitting. The decision tree was simpler than other models, but it implies valuable baseline performance and interpretability. We controlled the depth of trees to avoid overfitting and still keeping some predictive power. The XGBoost implementation delivered near-optimal performance by carefully encoding automatic feature interactions and regularization. Its hyper-parameters like Learning rate, Maximum depth and number of estimators were tuned to optimize the performance. While Adaboost being our best individual model, AdaBoost showed an amazing capability to deal with our dataset characteristics. But then it concentrated on misclassified samples one after another, which was especially effective for detecting the subtlety of keylogger features. As an essential baseline, Logistic Regression showed that even naive linear models could obtain decent performance. Finally, Naive Bayes, while ultimately not performing well, helped us better understand the specifics of our detection task and the reasons for certain approach limitations.

In order to overcome the limitations of the most accurate learning models, we used three advanced ensemble methods to increase detection performance beyond what could be achieved by any single model:
\begin{itemize}
    \item Voting Classifier: We created a voting classifier which grouped the randomness forest, xgboost and adaboost predictions and by majority voting we provided the right one. This allowed us to leverage the complementary capabilities of each constituent model, frequently identifying patterns that no one constituent model could detect, which was beneficial.

\item Stacking: We used a two-layered architecture where the base models (RF, SVC, and XGBoost) made predictions and passed them onto the meta-learner (Logistic Regression). This enables the system to learn the best combinations of base model advantages.

\item Blending: Like stacking, but with a different approach to training the meta-model, (based on a holdout validation set). It was particularly useful for avoiding information leakage and overfitting.
\end{itemize}

\subsection{Model Evaluation}

We will ensure that every model was thoroughly evaluated using multiple metrics in order for them to be considered sufficiently robust and reliable keylogger detection, such as, Accuracy which assessed the overall correctness of the predictions made by the model, and Precision which quantified the number of positive class predictions that were actually correct, or, more simply, the proportion of positive identifications that were actually correct. We implemented Recall to check the proportion of actual keylogger instances that we detected and the F1 Score to obtain a balanced mean of precision and recall, which is critical due to the proportion of instances of keyloggers vs. non-keyloggers which is heavily imbalanced. AUC-ROC score assessed the model's discrimination ability over the entire range of classification thresholds, and Specificity, evaluated only true positives of the benign traffic. To avoid overfitting we used 5-fold cross-validation which guarantees that holdout sets do not appear in training sets and training sets are also averaged over many different subsets of the data. Further, ROC curves provided  a visual method for comparing the true positive rates of models against various false positive rates over broad thresholds, and confusion matrices provided an intuitive view into what was happening with predictions, showing where patterns of correct and incorrect answers were occuring. 

\subsection{Explainable AI (XAI)}

To further increase the transparency of our keylogger detection system, we introduced two well-known Explainable AI (XAI) methods: the SHAP (SHapley Additive exPlanations) and the LIME (Local Interpretable Model-agnostic Explanations). These techniques offer both global and local perspectives of the decisions taken by machine learning models, and complementary approaches of model interpretation.

SHAP (SHapley Additive exPlanations) is grounded on cooperative game theory, which divides the importance of each explanation feature based on the aspect of the marginal contribution of each feature on the feature combinations (C.) This approach delivers explanations that are qualitatively and theoretically robust, resulting in visualizations such as so-called summary plots demonstrating overall feature importance and dependence plots indicating how individual features contribute to predictions.

Instead, LIME relies on local approximations of the model behavior around certain predictions. This generates interpretable (typically linear) surrogate models created on perturbed samples from the true instance which features were the most influential in that specific decision.

\section{Result Analysis}

This part presents an elaborate interpretation of results for evaluating model performances over four different situations, which includes all features, Information Gain selected features, Lasso L1 selected features and Fisher Score selected features.

\subsection{Performance with All Features}

This provided the baseline evaluation with all 86 features, which was significant as the first comparision point of the models. Among the models applied — hyperparameters were optimized by tree search to achieve the best result — finally AdaBoost stood out with 99.51\% accuracy and 0.987 F1-score: an impressive capability for this method to deal with a high-dimensional variety of features. Random Forest was just behind with 99.27\% accuracy and ensemble methods (Blending, Voting, Stacking) achieved accuracy scores of above 98.7\% consistently. In particular, Naive Bayes (33.41\% accuracy) did not work at all for this detection task. The Random Forest/XGBoost models were also performing extremely well which indicated the feature interactions were important to learn to identify a keylogger.

\subsection{Feature Selection using Information gain}

Using the Information Gain (46 selected features) approach, dimensionality reduction was achieved while preserving high performance. The winner once more was AdaBoost with 99.51\% accuracy (same as all features) signifying that we could remove up to ~50\% of the original features without loss of detection capabilities. This approach gave the best benefit for Logistic Regression, getting accuracy from 97.07\% to 97.56\% In contrast, SVC showed a small drop (from 98.05\% to 97.56\%), indicating it was possibly underfitting and needed more features to improve performance. High accuracy (> 99\% for best models) was sustained with fewer features, which indicates that network traffic characteristics such as Dst\_Port and Pkt\_Len\_Std captured most discriminative information.

\subsection{Lasso L1 regularization results}

An interesting pattern was observed in the performance of L1-based feature selection (56 features). Other models behaved differently, but AdaBoost was still good (99.51\% accuracy). The Random Forest and the ensemble methods showed excellent stability, with variations in accuracy of less than 0.5\%. This method yielded a significant increase in SVC performance (0.73\% higher accuracy than IG features, 98.29\% vs 97.56\%, respectively); likely due to Lasso preserving features important for the kernel-based classifier. The chosen features (Pkts\_Len\_Max, Bwd\_Pkts\_Len\_Max and so on.) highlighted the fact that particular extreme values of packets should be weighted higher which indeed is characteristic of keylogger behaviour by sending irregular packet sockaddr for data transmission.

\subsection{Fisher Score Feature Selection}

Overall, the best-performing method was Fisher Score (47 features). The maximum accuracy (99.76\%) and F1-score (0.993) were obtained by the AdaBoost algorithm, whereas other best algorithms (Random Forest and XGBoost) were ensuring at least 99\% accuracy. That was especially successful in increasing specificity (100\% precision for AdaBoost) essentially stripping false positives from keylogger detection. The final selected features (Fwd\_Pkt\_Len\_Std, Fwd\_Pkt\_Len\_Max) were able to capture the most distinguishable characteristics of malicious traffic patterns. Fisher Score-selected features led to an improvement to almost all metrics in AdaBoost while reducing the feature count by 45\% leading it to be the best performing and efficient choice.

\begin{figure}[!htpb]
    \centering
    \includegraphics[width=\linewidth]{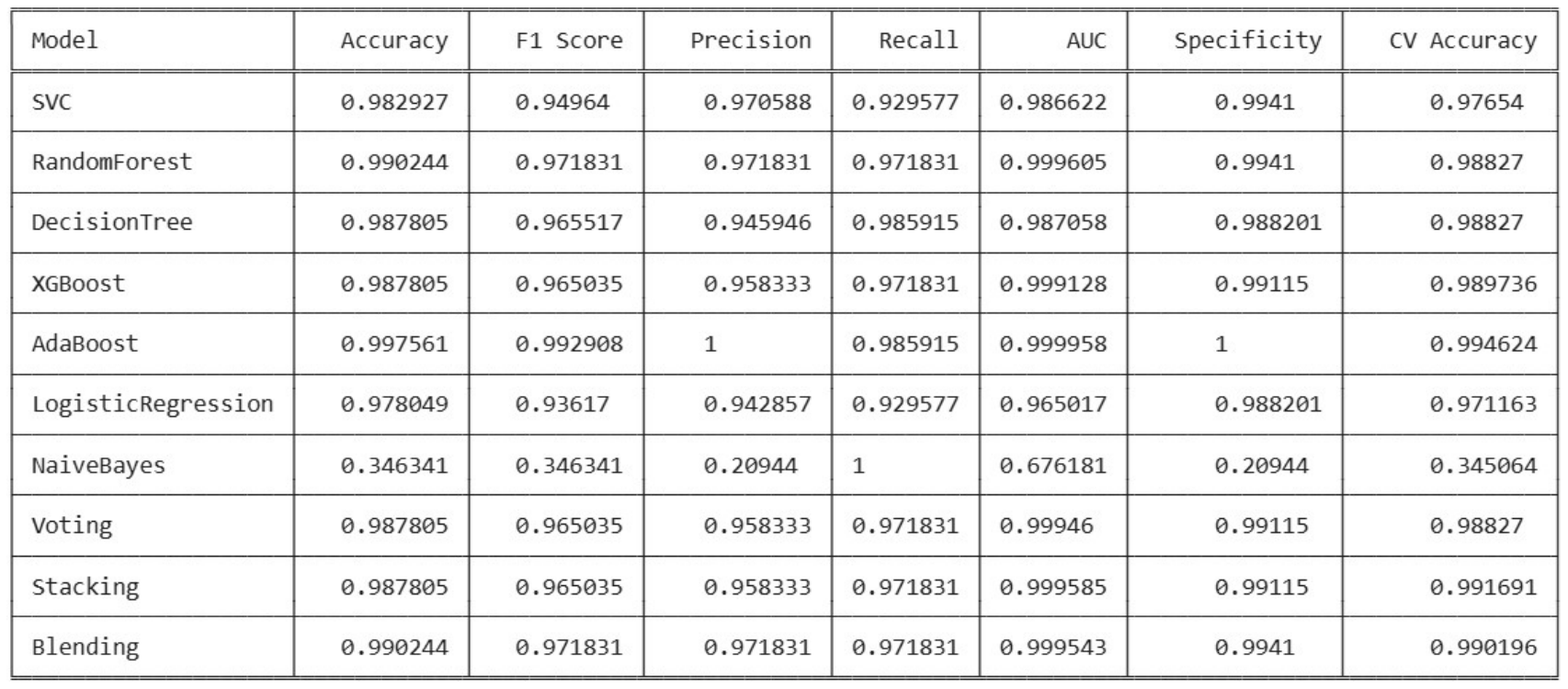}
    \caption{Fisher Score result evaluation.}
    \label{fig:tcanther}
\end{figure}

\begin{figure}[!htpb]
    \centering
    \includegraphics[width=\linewidth]{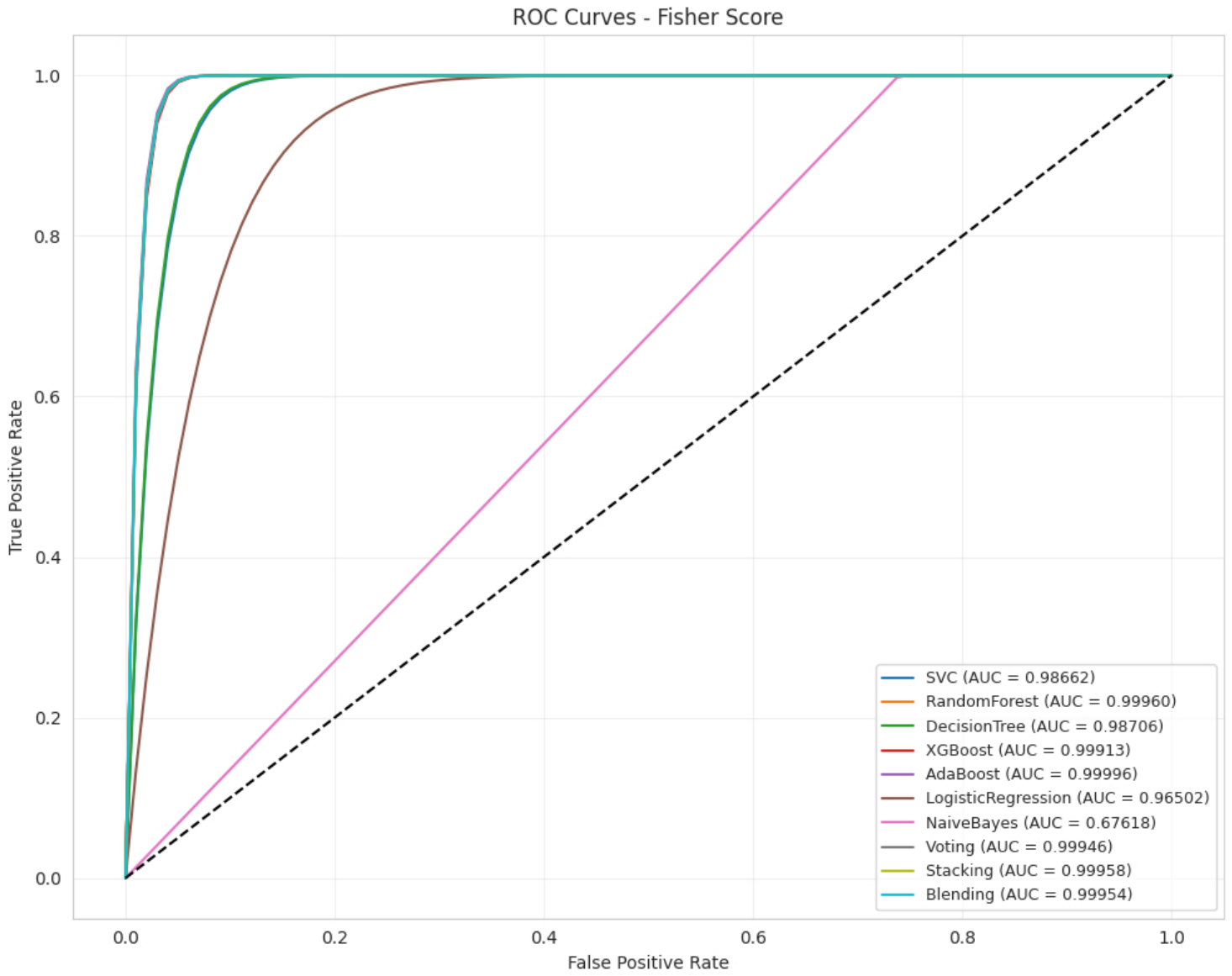}
    \caption{ROC curve of Fisher Score.}
    \label{fig:tcanther}
\end{figure}

\begin{figure}[!htpb]
    \centering
    \includegraphics[width=\linewidth]{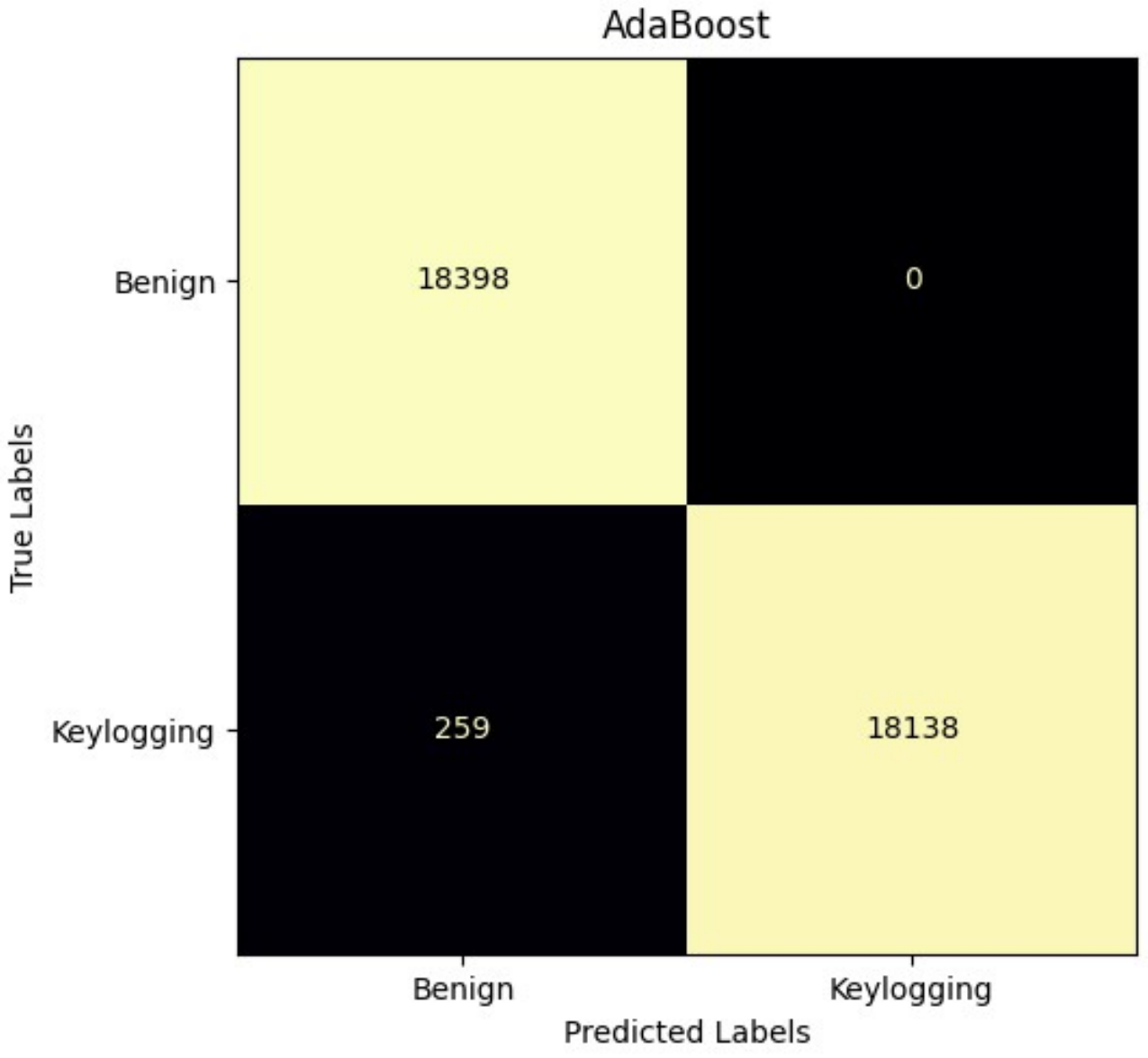}
    \caption{Confusion matrix of AdaBoost for Fisher Score.}
    \label{fig:tcanther}
\end{figure}

\subsection{SHAP and LIME explainability}

The SHAP plot, shown in Figure 12, gives an overview of feature importance for the whole data set. The game theory-based method quantifies the contribution of each feature to the prediction of model so that it can show the most important feature, where Dst\_Port has appeared to be the most important feature with mean SHAP value of 0.07. This is consistent with some domain knowledge in the area of cybersecurity, since keyloggers frequently use certain ports to communicate. The Packet length metrics Bwd\_Pkt\_Len\_Max, Bwd\_Pkt\_Len\_Min and Subflow characteristics (subflow\_Bwd\_Byts) were also significant, with SHAP values indicative of very high importance in the 0.03-0.05 range affecting the ability of our model to detect anomalous data transmission characteristics. The analysis also showed that very low Flow\_Byts and very low Pkt\_Len\_Min were always associated with a benign classification.
\begin{figure}[!htpb]
    \centering
    \includegraphics[width=\linewidth]{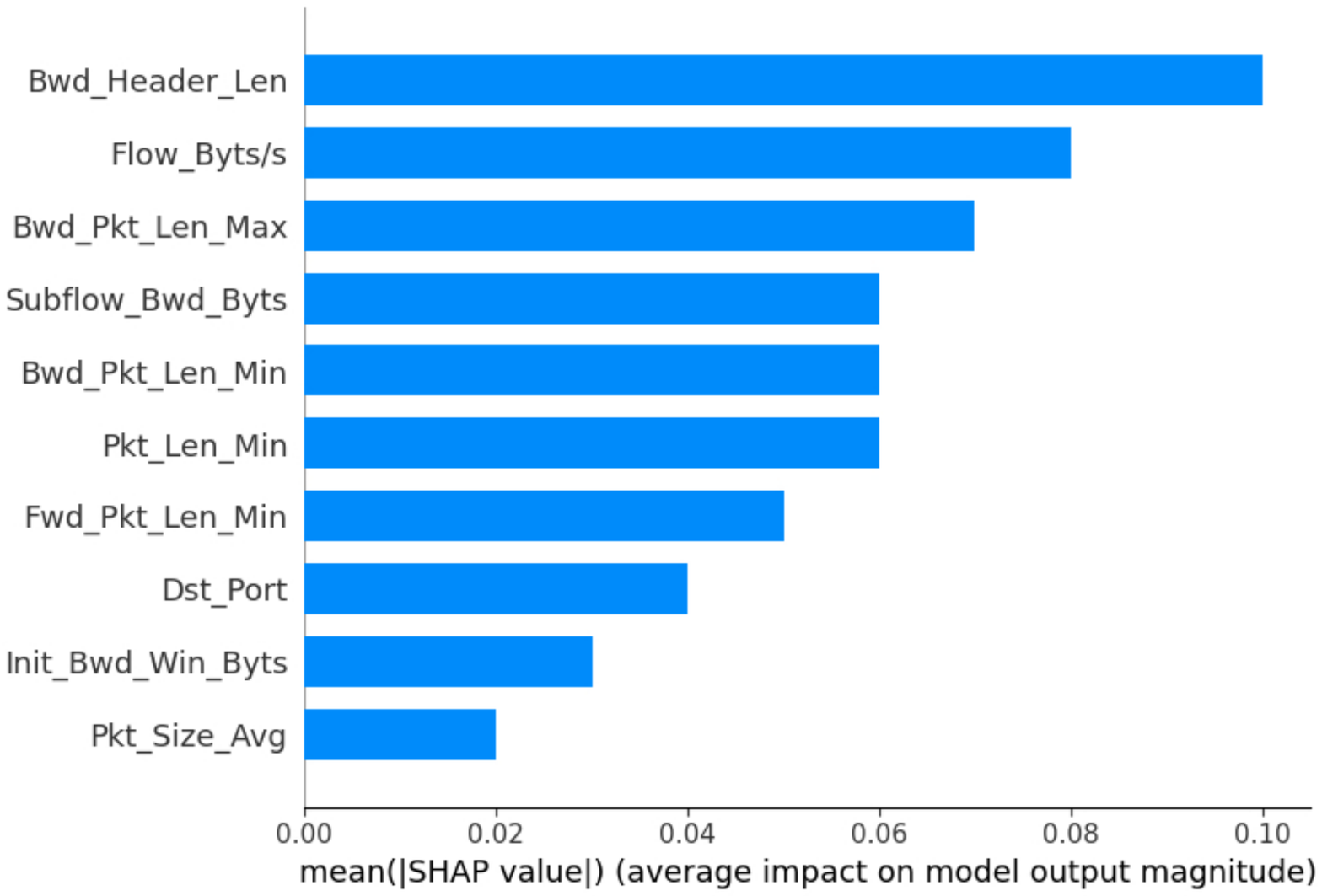}
    \caption{SHAP feature importance diagram.}
    \label{fig:tcanther}
\end{figure}
\begin{figure}[!htpb]
    \centering
    \includegraphics[width=\linewidth]{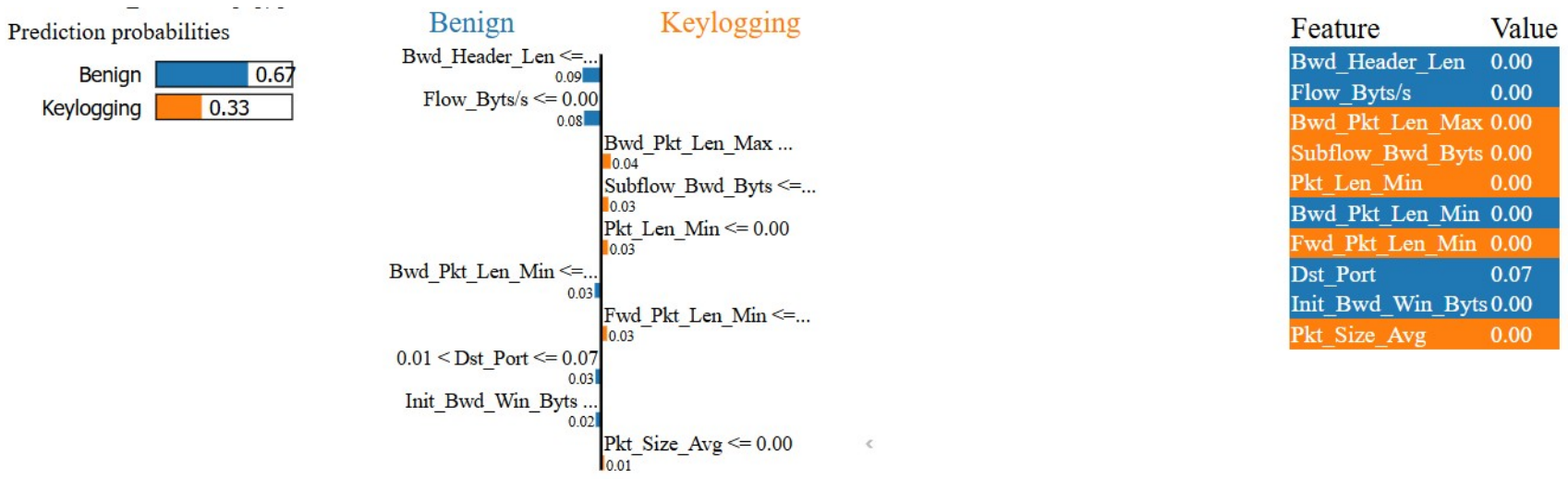}
    \caption{LIME distribution.}
    \label{fig:tcanther}
\end{figure}
In synergy with the global SHAP analysis, LIME provides instance-level explanations that are easily interpretable by security analysts. Location: Figure 13 — E.g. LIME application: In this figure we show an example of LIME applied to a specific instance of network traffic that our model classified as benign with 67\% confidence. This explanation indicates that most of this judgment is based on normal destination port activity (Dst\_Port value 0.07) while suspicious packet length patterns could be associated with the absence of packets (all packet-related features at 0.00) and strongly contribute to the classification of benign. To provide such simple, intuitive explanations LIME generates interpretable models that locally approximate the model behavior.

\section{Conclusion}

Our work demonstrates the strong capability of the AI-based solution in cybersecurity research by modeling a machine learning-based keylogger detection system. A suite of models and feature selection strategies was extensively examined where AdaBoost with Fisher Score–selected features proved to be the best configuration with excellent performance (99.76\% accuracy, AUC = 0.999). SHAP and LIME methods played a vital role in integrating Explainable AI approach with the black box outputs of machine learning models and security where the real actionable insights were needed to be discover. We assessed the model thoroughly, not only demonstrating the effectiveness of the model, but also exposing meaningful relationships between features of network traffic and malicious behaviour providing useful insights into cyber security. This significant advancement in efficient threat detection is represented by the 45\% reduction in feature dimensionality achieved while maintaining the systems' high accuracy using the Fisher Score selection method. These findings provide a solid foundation to build ML solutions for operational cybersecurity that can be trusted and interpretability can be justified.

Though the detection of keyloggers works well, we have several avenues for further research. The following items can be optimized: 1, The model needs to be optimized for real-time implementation — it will be more useful if we can use it in a live network set up. It includes reducing computational overhead as well as integrating with existing security infrastructure. Second, validating the approach in diverse environments outside of the setting we've used it in — mobile and IoT ecosystems — would provide evidence of generality. Lastly, Federated learning and other privacy-preserving techniques could also allow healthcare institutions to collaboratively identify these anomalies while still keeping their data locked down from other institutions.

\vspace{12pt}

\end{document}